\documentclass[journal,twoside,web]{ieeecolor}
\usepackage{tmi}
\usepackage{cite}
\usepackage{amsmath,amssymb,amsfonts}
\usepackage{algorithmic}
\usepackage{graphicx}
\usepackage{textcomp}
\usepackage{comment}
\usepackage{multirow}
\usepackage[table]{xcolor}
\usepackage{booktabs}
\usepackage[normalem]{ulem}

\def\BibTeX{{\rm B\kern-.05em{\sc i\kern-.025em b}\kern-.08em
    T\kern-.1667em\lower.7ex\hbox{E}\kern-.125emX}}
\begin{document}
\title{ReCo-KD: Region- and Context-Aware Knowledge Distillation for Efficient 3D Medical Image Segmentation}
\author{Qizhen Lan, Yu-Chun Hsu, Nida Saddaf Khan, and Xiaoqian Jiang, \IEEEmembership{Member, IEEE}
\thanks{
\raggedright This work was supported in part by the U.S. National Institutes of Health (NIH) under Grant No. R01AG066749, R01AG066749-03S1, and R01AG082721. (Corresponding author: Xiaoqian Jiang.)}
\thanks{\raggedright Qizhen Lan, Yu-Chun Hsu, Nida Saddaf Khan, and Xiaoqian Jiang are with the McWilliams School of Biomedical Informatics, The University of Texas Health Science Center at Houston (UTHealth Houston), Houston, TX 77030, USA (e-mail: qizhen.lan@uth.tmc.edu; yu-chun.hsu@uth.tmc.edu; nida.s.khan@uth.tmc.edu; xiaoqian.jiang@uth.tmc.edu).}
} 
\maketitle

\begin{abstract}

Accurate 3D medical image segmentation is vital for diagnosis and treatment planning, but state-of-the-art models are often too large for clinics with limited computing resources. Lightweight architectures typically suffer significant performance loss. To address these deployment and speed constraints, we propose Region- and Context-aware Knowledge Distillation (ReCo-KD), a training-only framework that transfers both fine-grained anatomical detail and long-range contextual information from a high-capacity teacher to a compact student network. The framework integrates Multi-Scale Structure-Aware Region Distillation (MS-SARD), which applies class-aware masks and scale-normalized weighting to emphasize small but clinically important regions, and Multi-Scale Context Alignment (MS-CA), which aligns teacher–student affinity patterns across feature levels. Implemented on nnU-Net in a backbone-agnostic manner, ReCo-KD requires no custom student design and is easily adapted to other architectures. Experiments on multiple public 3D medical segmentation datasets and a challenging aggregated dataset show that the distilled lightweight model attains accuracy close to the teacher while markedly reducing parameters and inference latency, underscoring its practicality for clinical deployment.

\end{abstract}


\section{Introduction}
\label{sec:intro}

In recent years, deep learning has significantly advanced 3D medical image segmentation, enabling precise delineation of complex anatomical structures. Convolution-based U-Net architectures have consistently demonstrated strong performance in this domain, particularly due to their ability to capture local, fine-grained contextual details—critical for tasks such as small tumor delineation and boundary identification. Transformer-based methods (e.g., SwinUNETR~\cite{hatamizadeh2021swin}, nnFormer~\cite{zhou2021nnformer}) target long-range dependencies but incur higher compute/memory and often do not yield consistent accuracy gains over CNNs in medical settings. A notable exception is MedNeXt~\cite{roy2023mednext}, which modernizes CNN design by incorporating transformer-inspired principles such as large receptive fields, reparameterizable blocks, and compound scaling. These enhancements allow MedNeXt to bridge the benefits of local-detail sensitivity and global context modeling. However, despite improved modeling capacity, MedNeXt remains compute- and memory-intensive, limiting deployment on CPU-only, mobile, or point-of-care systems. Consequently, clinical adoption is often constrained more by efficiency than by algorithmic accuracy.

\begin{figure}[t]
    \centering
    \includegraphics[width=0.9\linewidth]{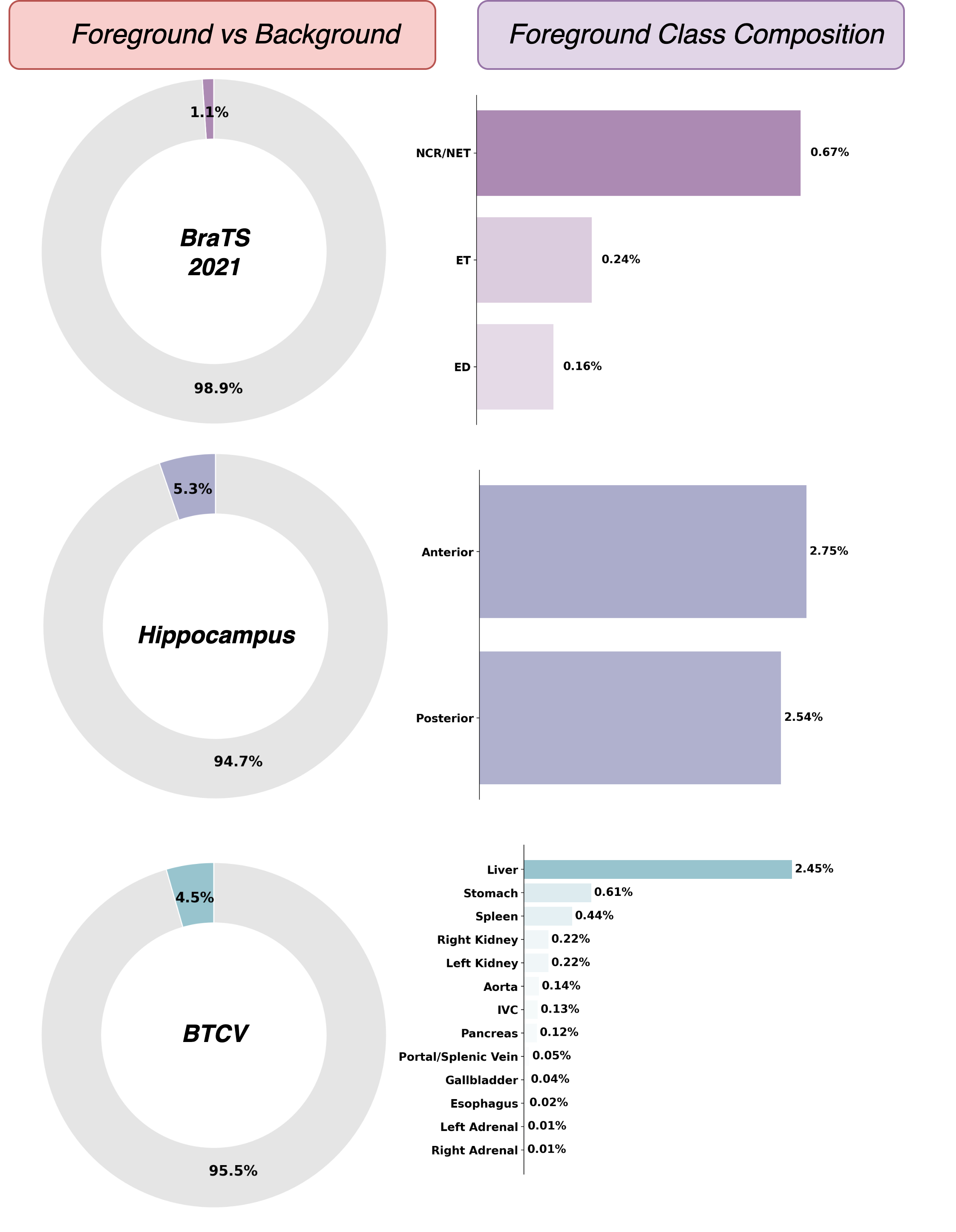}
    \caption{Voxel distribution across background and foreground classes in three medical segmentation datasets. The left donut charts show the dominance of background voxels—BraTS 2021: 98.9\%, Hippocampus: 94.7\%, BTCV: 95.5\%. The right bar charts reveal strong foreground imbalance, e.g. BTCV (Liver 2.45\%; most other organs <1\%). This imbalance suggests that equal voxel weighting in knowledge distillation may overlook small yet clinically critical structures.}
    \label{fig:voxel_distribution}
\end{figure}


\begin{figure}[t]
    \centering
    \includegraphics[width=0.95\linewidth]{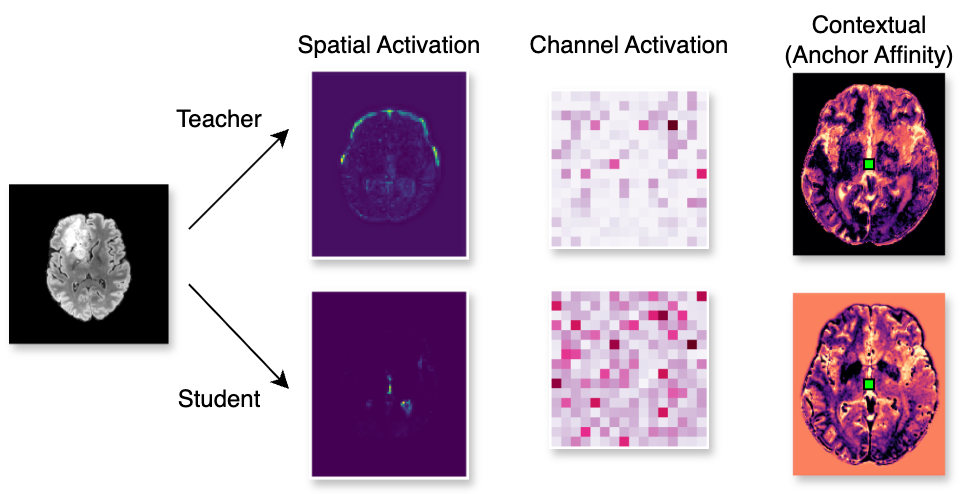}
    \caption{
    Visualization of spatial, channel, and contextual representations from the teacher and the student (encoder stage 1). The green square highlights the anchor point, and the surrounding maps show responses relative to this position. The discrepancies across all levels indicate the representational gap, motivating the need for knowledge distillation to align them.
    }
    \label{fig:teacher_student_diff}
\end{figure}

In response to these deployment challenges, recent research has proposed a range of efficient architectures for 3D medical image segmentation, aiming to reduce model size and computational cost to speed up inference. For example, ENet reduces model complexity through early downsampling and the use of dilated convolutions~\cite{paszke2016enet}, while ERFNet employs factorized convolutions and a residual architecture to maintain efficiency without sacrificing performance~\cite{romera2017erfnet}. MobileNetV3 leverages neural architecture search (NAS), depthwise separable convolutions, and hardware-friendly activation functions to optimize both latency and accuracy on mobile devices~\cite{howard2019searching}. In the medical domain, Mobile-UNet~\cite{jing2022mobile}, UNETR++~\cite{shaker2024unetr++}, SegFormer3D~\cite{perera2024segformer3d}, and SlimUNETR~\cite{pang2023slim} introduce efficiency-focused architectures, but often at the expense of segmentation accuracy.

To bridge this performance–efficiency gap, knowledge distillation (KD)~\cite{hinton2015distilling} has been widely explored in computer vision, transferring semantic knowledge from large teacher models to compact student models. While KD initially focused on classification, subsequent work extended it to semantic segmentation by transferring spatial structures, pairwise feature relations~\cite{liu2023structured}, intra-class pixel variations~\cite{wang2020intra}, and cross-image semantic affinities~\cite{yang2022cross}.  In medical image segmentation, KD remains underexplored—particularly for 3D volumes—despite early efforts on boundary-guided distillation~\cite{wen2021towards}, region affinity modules~\cite{qin2021efficient}, and structured feature filtering~\cite{zhao2023mskd} for 2D images. Encoder-level feature transfer in 3D segmentation has been attempted only in a few works~\cite{wen2021towards,zhao2023mskd}, and none explicitly address the joint challenges of spatial imbalance and global contextual coherence.

A fundamental obstacle is the region- and scale-imbalance in volumetric data: small yet critical structures (e.g., adrenal glands, hippocampal subfields, enhancing tumor) occupy less than 1\% voxels, whereas large organs/background dominate the volume. As shown in Fig. \ref{fig:voxel_distribution}, more than half of the anatomical classes in BTCV, Hippocampus, and BraTS2021 datasets have a volume ratio below $0.5\%$, with the largest-to-smallest class volume ratio exceeding 200:1. This extreme imbalance biases distillation objectives toward dominant structures, suppressing supervision signals from rare but critical regions and leading to suboptimal generalization. Similar issues have been investigated in object detection \cite{zhao2022decoupled, lan2025acam, lan2024gradient}, where region-aware distillation strategies (e.g., spatial masks, multi-scale feature alignment) have shown benefits. However, direct adaptation to 3D medical segmentation is non-trivial due to the high spatial resolution, dense voxel dependencies, and memory constraints of volumetric networks. 


Beyond voxel imbalance, preserving contextual coherence is equally critical. As illustrated in Fig.~\ref{fig:teacher_student_diff}, knowledge mismatch manifests not only at the spatial level, but also in channel activations and contextual dependencies. Without explicitly modeling these differences, the student may inherit only partial or noisy guidance, limiting its ability to capture both fine-grained details and long-range anatomical coherence.
Accurate 3D segmentation requires modeling inter-voxel dependencies to maintain consistent anatomical relationships across organs and subregions. Most KD methods either ignore such global relationships or apply them only at the final output level, missing the opportunity to guide the student’s intermediate feature hierarchy toward anatomically plausible representations. 

To address these limitations, we propose \textbf{ReCo-KD} (Region-and-Context-aware Knowledge Distillation), a unified framework that jointly enforces \emph{multi-scale structure-aware region distillation} (MS-SARD) and \emph{multi-scale contextual alignment} (MS-CA) for 3D medical segmentation. MS-SARD employs class-aware masks, scale-normalized weighting, and attention-enhanced feature matching to emphasize semantically critical but under-represented voxels. MS-CA aligns teacher–student relational feature across multiple scales to preserve long-range anatomical dependencies and maintain contextual consistency. The framework operates exclusively on intermediate representations during training, introducing \emph{no additional inference cost}, and is fully compatible with state-of-the-art segmentation backbones.

In summary, our main contributions are as follows:

\begin{itemize}
    \item We propose a Multi-Scale Structure-Aware Region Distillation (MS-SARD) module that highlights semantically critical voxels through class-aware masking, scale normalization, and spatial–channel response guidance across encoder stages.
    \item We introduce a Multi-Scale Context Alignment (MS-CA) module that transfers long-range dependencies by aligning teacher–student affinity structures at multiple feature levels.
    \item Our framework is implemented on top of nnU-Net, providing plug-and-play integration with automatic configuration and support for multi-modal inputs.
    \item The method delivers substantial computational savings—reducing parameters and FLOPs while preserving near-teacher accuracy—and is extensively evaluated on multiple public 3D segmentation datasets and a challenging aggregated dataset, demonstrating strong performance.
\end{itemize}

\section{Related Works}

\subsection{3D Medical Image Segmentation}
Semantic segmentation plays a crucial role in medical image analysis as it enables precise delineation of anatomical structures. Since the introduction of the encoder–decoder framework in U-Net \cite{ronneberger2015u}, convolutional neural networks (CNNs) have demonstrated strong performance in 3D medical image segmentation~\cite{cciccek20163d, isensee2021nnu, zhou2019unet++}. Variants such as UNet++ \cite{zhou2019unet++} improve multi-scale feature representation through redesigned skip connections, while CPF-Net \cite{feng2020cpfnet} leverages context pyramid fusion to capture global and multi-scale contextual information. While transformer-based architectures have demonstrated impressive capabilities in capturing long-range dependencies via self-attention~\cite{vaswani2017attention}. 
TransUNet~\cite{chen2021transunet} combines transformer-based global context modeling with CNN-based localization for improved segmentation accuracy. Swin UNETR \cite{hatamizadeh2021swin} further reduces computational costs with hierarchical windowed attention \cite{liu2021swin}, but the overhead from window shifting and deep transformer layers remains substantial. Despite promising accuracy, volumetric attention and deep decoders substantially increase FLOPs and memory, hindering deployment on resource-limited clinical hardware. This motivates compact CNN-centric frameworks and targeted optimization to balance performance and efficiency in 3D settings.

\vspace{-0.6em}
\subsection{Lightweight U-Net for Medical Segmentation}
To support segmentation on resource-constrained platforms, many lightweight designs pursue large efficiency gains while accepting small accuracy trade-offs. Techniques from general computer vision, such as factorized convolutions and depthwise separable convolutions \cite{chollet2017xception}, have been incorporated into architectures like MobileNet \cite{sandler2018mobilenetv2}, ShuffleNet \cite{zhang2018shufflenet}, and EfficientNet \cite{tan2019efficientnet}. In medical imaging, Mobile-UNet–style variants (e.g., MobileUNetV3  \cite{jing2022mobile} ) embed MobileNetV3 \cite{howard2019searching} encoders into U-Net, reporting substantial FLOP cuts with competitive accuracy. UNeXt \cite{valanarasu2022unext} replaces some convolutional blocks with tokenized-MLP modules to balance local detail and global context while remaining efficient. 3D efficiency-oriented models such as UNETR++ \cite{shaker2024unetr++}, SegFormer3D \cite{perera2024segformer3d}, and SlimUNETR \cite{pang2023slim} further reduce computation, but often at the cost of lower segmentation accuracy—particularly in multi-organ or small-structure segmentation. Despite lower FLOPs, lightweight backbones tend to underperform on rare/small anatomies and fine boundaries, reflecting weak supervision for under-represented regions and limited global coherence. We therefore complement lightweight design with an architecture-agnostic distillation scheme that up-weights small/rare regions and preserves global context during training, adding zero inference-time cost.
\vspace{-0.6em}
\subsection{Knowledge Distillation for Medical Segmentation}
Knowledge distillation (KD) was originally introduced for image classification \cite{hinton2015distilling}, where a student network learns from a teacher network’s softened outputs. In dense prediction tasks such as semantic segmentation, KD has been extended to include feature-based distillation, structural relation transfer, and class-wise affinity modeling \cite{liu2023structured,wang2020intra,yang2022cross}. 

In medical image segmentation, KD approaches have explored boundary-guided distillation \cite{wen2021towards}, region-wise feature transfer \cite{qin2021efficient}, and multi-scale structured distillation \cite{zhao2023mskd}. Although effective, many methods target 2D settings and under-address 3D challenges: class/region imbalance biases learning toward dominant/background voxels, and insufficient multi-scale relational alignment can yield anatomically fragmented student predictions.
We propose a region- and context-aware KD that (i) re-weights supervision using class-aware masks and scale-normalized factors to emphasize rare yet critical voxels, and (ii) aligns teacher–student inter-voxel relations across multiple feature levels to preserve global anatomical consistency—all training-only, with no inference-time overhead.

\section{METHODOLOGY}
\begin{figure*}[t]
    \centering
    \includegraphics[width=\linewidth]{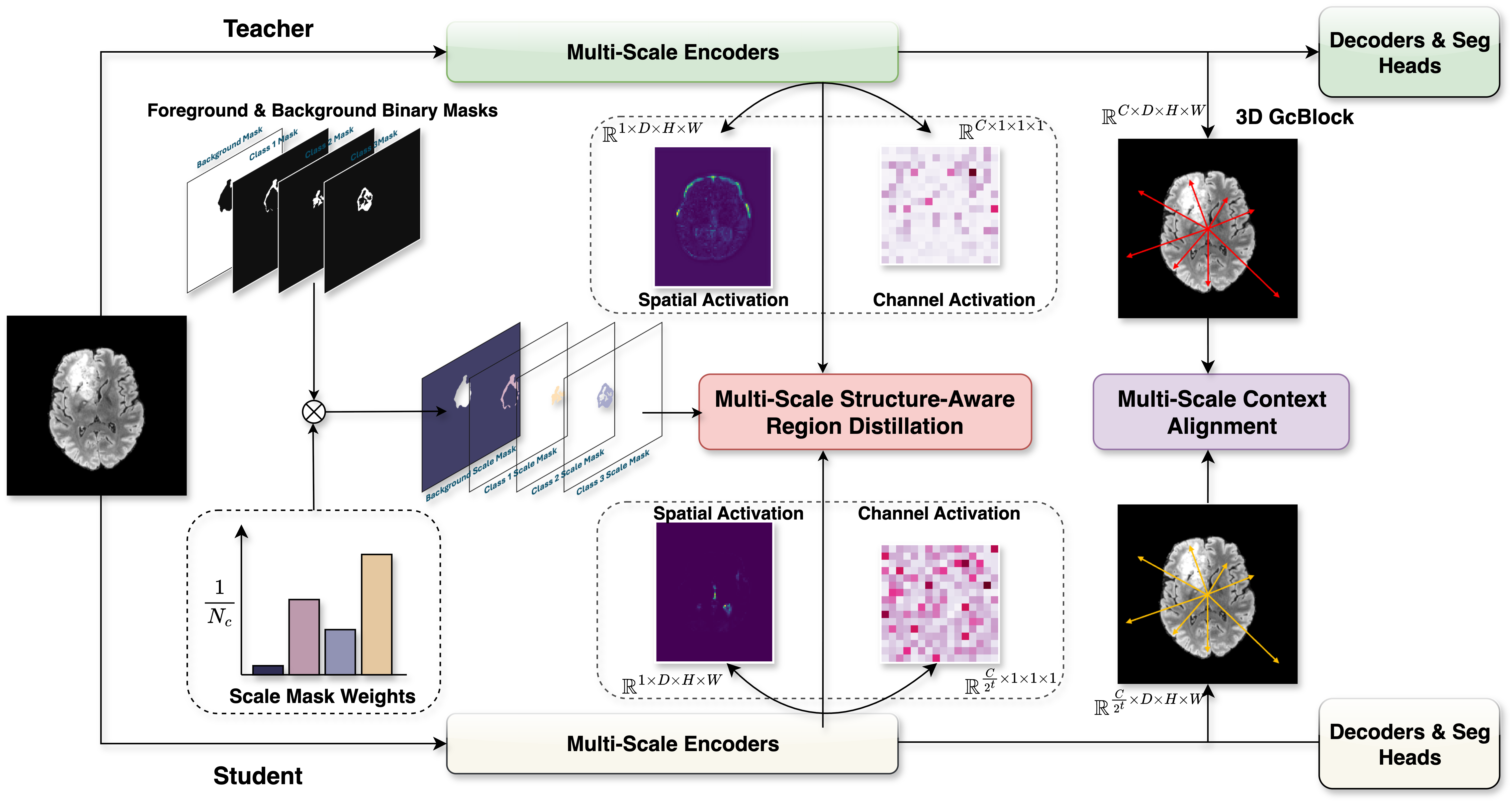}
    \caption{
    Overview of the proposed Region- and Context-aware Knowledge Distillation (ReCo-KD) framework for 3D medical image segmentation. 
    Teacher and student share the same backbone, with the student using a channel-reduced width (\(C/2^{t}\)). 
    Multi-scale feature maps from all encoder stages are distilled by two complementary modules: 
    \textbf{Multi-Scale Structure-Aware Region Distillation (MS-SARD)}, which highlights class-specific regions with scale-normalized weighting, 
    and \textbf{Multi-Scale Context Alignment (MS-CA)}, which aligns teacher–student affinity patterns to transfer long-range dependencies. 
    Together these modules enable the compact student to achieve near-teacher accuracy with greatly reduced computation.
    }
    \label{fig:overview}
\end{figure*}


Our method builds on nnU-Net \cite{isensee2021nnu}, a self-configuring 3D segmentation framework that adapts preprocessing, architecture, and training to each dataset with minimal manual tuning. For resource-constrained deployment, we construct a compact student by uniformly scaling the channel number across blocks while keeping depth, strides, skip connections, and input resolution unchanged.
We instantiate a family of students parameterized by $t\in \{0,1,2,3\}$ with reduction multiplier $2^{-t}$ (i.e., \(t{=}0\!\Rightarrow\!\times1\), \(t{=}1\!\Rightarrow\!\times\frac{1}{2}\), \(t{=}2\!\Rightarrow\!\times\frac{1}{4}\), \(t{=}3\!\Rightarrow\!\times\frac{1}{8}\)). 
For each stage \(i\) with base channels \(C_i\), we set the student channels as:

\begin{equation}
    C_i' \;=\; \max(C_{\min},\ 2^{-t}\,C_i),
\end{equation}

\noindent where the minimum value $C_{\min}$ is set to 4 to preserve representation capacity under high compression rate. 

Using a different lightweight backbone (e.g., ShuffleNet \cite{zhang2018shufflenet}, MobileNetV3 \cite{jing2022mobile}) requires non-trivial integration and may break nnU-Net’s auto-configuration. Uniform width scaling is simpler and fully compatible with nnU-Net. It offers (i) minimal engineering across datasets and (ii) seamless reuse of the pipeline.
A known drawback is that aggressive channel reduction can weaken discrimination, especially for small structures and low-contrast regions. Our distillation method is designed to counter this effect.

We apply distillation on the encoder outputs at multiple stages. A common feature-based objective is:
\vspace{-0.3em}
\begin{equation}
\mathcal{L}_{\text{feat}} = \sum_{c=1}^{C} \sum_{i=1}^{D} \sum_{j=1}^{H} \sum_{k=1}^{W}\left( F^T_{c,i,j,k} - f(F^S_{c,i,j,k}) \right)^2 \\
\end{equation}

\noindent where the $F^T$ and $F^S$ indicate the feature from the teacher and student network, respectively. $f(\cdot)$ is an alignment function (e.g., a $1 \times 1 \times 1$ convolution layer) to reshape the student feature to match the teacher's dimension. $C, D, H$, $W$ denote the channel, depth, height, and width of the feature, respectively. This uniform loss treats all voxels equally. It ignores class/region imbalance and long-range context, which are crucial in 3D medical images. To address this, we introduce a dual-branch KD with \emph{multi-scale structure-aware region distillation}, which up-weights semantically important but under-represented voxels across scales, and \emph{multi-scale contextual alignment}, which aligns long-range dependencies to maintain global anatomical consistency between teacher and student. Importantly, all components are training-only, so inference-time complexity matches the lightweight student baseline. All components operate only during training, so inference complexity remains identical to the lightweight student. As shown in Figure \ref{fig:overview}, the training pipeline includes both the full teacher and the compact student, and distillation is applied at multiple encoder levels.

\subsection{Multi-Scale Structure-Aware Region Distillation (MS-SARD)}
\label{sec:MS-SARD}

To address the problem of region and scale imbalance, the first branch of ReCo-KD focuses on transferring fine-grained structural knowledge from the teacher to the student across multiple encoder stages. We proposed Multi-Scale Structure-Aware Region Distillation (MS-SARD), which derives its supervision from class-aware structural masks combined with scale-normalized voxel weighting, enabling the student to focus proportionally on small and clinically important structures.

Firstly, for each anatomical class region $r \in \{0, 1, 2, ..., \mathcal{R}\}$, a binary region mask $M^{r}$ isolates the voxels belonging to class region $r$ and all others (including background class region). The region mask is defined as:

\begin{equation}
M_{i,j,k}^{r} = 
\begin{cases}
1, & \text{if } (i, j, k) \in \Omega_r, \\ 
0, & \text{otherwise},
\end{cases}
\end{equation}

\noindent where $\Omega_r$ denotes the set of voxels that belong to class region $r$, and $(i,j,k)$ indexes a spatial location in the 3D volume. If $(i,j,k)$ falls in the corresponding class region, then $M_{i,j,k} = 1$, otherwise it is 0.

While region masking enforces semantic focus, the severe voxel imbalance across anatomical structures still biases the distillation objective toward large-volume classes (particularly the background), because these regions contain substantially more voxels and thus dominate the loss. To mitigate this, we introduce a class-wise scale mask that rebalances supervision according to class size. We define a class-wise scale mask $S^{r}$ for each class region $r$ as follows:

\begin{equation}
S_{i,j,k}^{r} = \frac{1}{N_r}, \quad \text{if } (i,j,k) \in \Omega_r,
\end{equation}

\noindent where $N_r$ is the total voxel count of the class region $r$:

\begin{equation}
N_r = \sum_{i=1}^{D} \sum_{j=1}^{H} \sum_{k=1}^{W} \mathbf{1} \left[(i,j,k) \in \Omega_r \right],
\end{equation}

\noindent if a voxel belongs to multiple classes, we assign it to the class with the largest weight $\frac{1}{N_r}$ (i.e., the smallest spatial coverage) when computing $S$. In this formulation, large-volume regions (e.g., background) receive smaller weights, while small but clinically critical structures receive larger weights. This class-aware rebalancing alleviates the dominance of background voxels and encourages the distillation process to place greater emphasis on under-represented regions. Consequently, the proposed scale mask $S^r$ serves as a normalization mechanism that equalizes the contribution of each anatomical structure, ensuring that supervision signals are more uniformly distributed across regions.

Furthermore, to highlight the most informative voxels and channels during distillation, we compute spatial and channel-wise activation masks from the feature map $F \in \mathbb{R}^{C \times D \times H \times W}$. We first compute aggregated activation statistics across channel and spatial dimensions:

\begin{equation}
A^{S}(F) = \frac{1}{C} \sum_{c=1}^{C} \left| F_c \right| ,
\end{equation}

\begin{equation}
A^{C}(F) = \frac{1}{DHW} \sum_{i=1}^{D} \sum_{j=1}^{H} \sum_{k=1}^{W} \left| F_{i,j,k} \right|,
\end{equation}

\noindent where $A^S(F) \in \mathbb{R}^{D \times H \times W}$ and $A^C(F) \in \mathbb{R}^{C}$ denote the intermediate spatial and channel activations. To obtain the final weighting masks, we normalize these statistics using temperature-controlled exponential scaling:
\begin{equation}
V^S(F) = DHW \cdot \frac{\exp \!\big( A^S(F) / T \big)}{\sum_{i,j,k} \exp \!\big( A^S_{i,j,k}(F) / T \big)},
\end{equation}

\begin{equation}
V^C(F) = C \cdot \frac{\exp \!\big( A^C(F) / T \big)}{\sum_{c} \exp \!\big( A^C_c(F) / T \big)} .
\end{equation}

\noindent where $T$ is a temperature parameter (proposed by \cite{hinton2015distilling}) that regulates the sharpness of the distribution; smaller values yield more peaked attention. These masks $V^S$ and $V^C$ reweight the distillation process to emphasize informative voxels and channels.

There exist inherent differences between teacher and student feature representations (see Fig.~\ref{fig:teacher_student_diff}). To bridge this gap during training, we employ teacher-derived masks to guide the student in both spatial and channel dimensions. To encourage the student to replicate the teacher’s activation patterns, we define an activation consistency loss as:
\vspace{-0.3em}
\begin{equation}
\mathcal{L}_{\text{ac}} = \gamma \cdot \left( \lVert V^S_t - V^S_s \rVert_1 + \lVert V^C_t - V^C_s \rVert_1 \right),
\end{equation}

\noindent where $t$ and $s$ denote the teacher and student, respectively, $\lVert \cdot \rVert_1$ is the L1 norm, and $\gamma$ is a balancing coefficient.

Beyond activation consistency, we introduce a structure-aware region distillation loss to directly align feature representations under the guidance of teacher-derived masks and all other masks. Specifically, we employ three types of masks: the binary region mask $M^r$, the class-aware scale mask $S^r$, and the spatial and channel activation masks $V^S$ and $V^C$. The structure-aware region distillation loss $L_{\text{sard}}$ at each stage is formulated as:
\begin{equation}
\begin{aligned}
\mathcal{L}_{\text{sard}} =\ & \sum_{r=1}^{\mathcal{R}} \sum_{c=1}^{C} \sum_{i=1}^{D} \sum_{j=1}^{H} \sum_{k=1}^{W} 
M_{i,j,k}^{r} \cdot S_{i,j,k}^{r} \cdot V^{S}_{i,j,k} \cdot V^{C}_c \\
& \cdot \left( F^T_{c,i,j,k} - f(F^S_{c,i,j,k}) \right)^2 ,
\end{aligned}
\end{equation}

\noindent where $F^{T}$ and $F^{S}$ denote the teacher and student feature maps, respectively, and $f(\cdot)$ is a lightweight convolutional projection layer that aligns the channel dimensions.  
The loss is weighted by the activation masks ($V^S$, $V^C$) together with the region and scale masks $M^r$ and $S^r$, providing balanced supervision and highlighting informative voxels and channels. This design allows the student not only to capture local semantic cues but also to retain discriminative information in both region-specific and context-aware manners.

The overall distillation objective is formulated as the Multi-Scale Structure-Aware Region Distillation (MS-SARD) loss, which integrates both feature alignment and attention alignment across all encoder stages:

\begin{equation}
\mathcal{L}_{\text{MS-SARD}} = \sum_{l=1}^{L} \left( L_{\text{sard}}^{(l)} + L_{\text{ac}}^{(l)} \right),
\end{equation}

where $L$ is the number of encoder stages for distillation. Each stage contributes $L_{\text{sard}}^{(l)}$ and $L_{\text{ac}}^{(l)}$, which together align region-specific features and activation patterns.

\subsection{Multi-Scale Contextual Alignment (MS-CA)}
\label{sec:context}

\begin{figure}[t]
    \centering
    \includegraphics[width=0.9\linewidth]{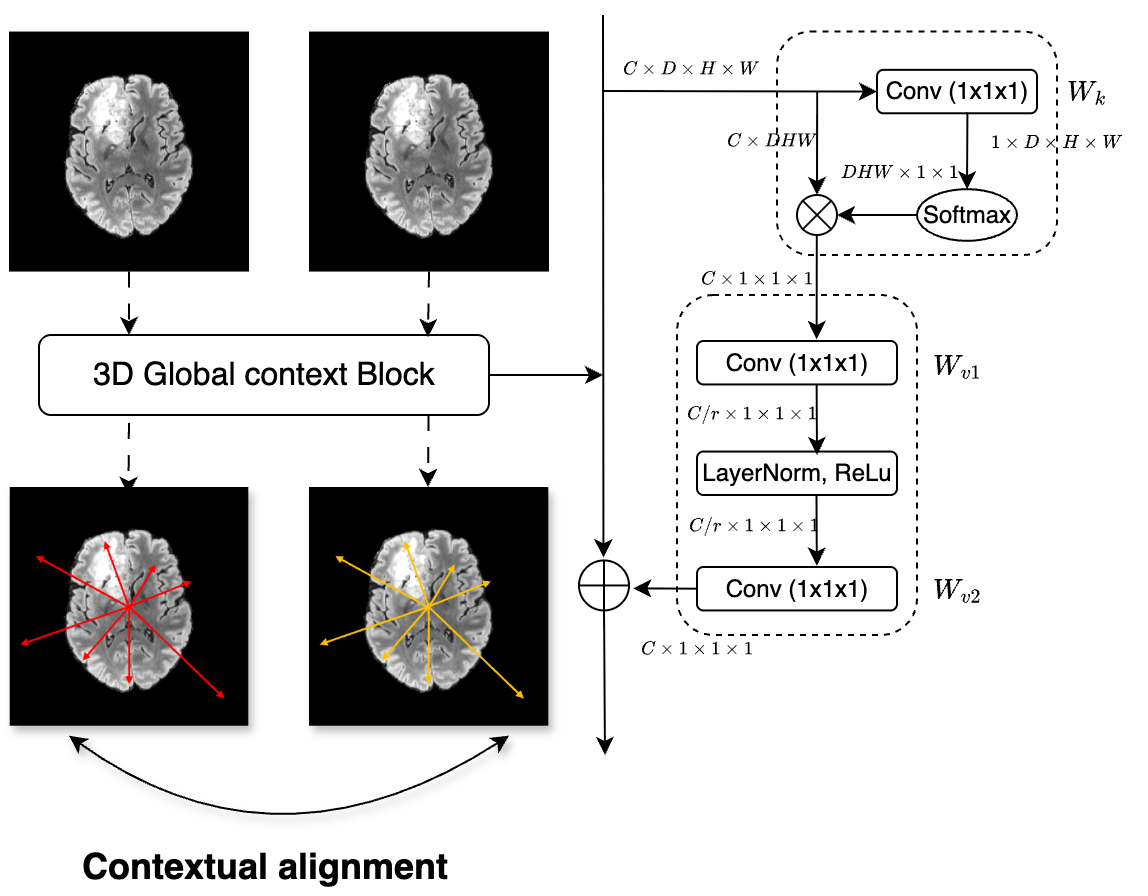}
    \caption{
    Illustration of contextual alignment distillation using the 3D Global Context Block. Feature volumes from both teacher and student encoder are taken as inputs to align contextual representations across stages.}
    \label{fig:3d_gcblock}
\end{figure}

While MS-SARD emphasizes class-discriminative cues, it may underrepresent global dependencies across the 3D volume. In medical segmentation, long-range interactions—such as those between anatomically related yet spatially distant structures—are crucial for structural completeness and global consistency. To address this limitation, we introduce Multi-Scale Contextual Alignment (MS-CA), which transfers holistic contextual patterns from teacher to student via a lightweight 3D global-context operator adapted from GC-blocks \cite{cao2019gcnet}. As depicted in Fig.~\ref{fig:3d_gcblock}, our method integrates a compact context-modeling module that distills holistic structural knowledge without interfering with the localized supervision provided by MS-SARD. By jointly leveraging localized and contextual guidance, the student is encouraged to produce anatomically coherent and semantically rich segmentation results.

Formally, the multi-scale contextual alignment loss $\mathcal{L}_{\text{MS-CA}}$ is defined as

\begin{equation}
\mathcal{L}_{\text{MS-CA}} = \lambda \cdot 
\sum_{l=1}^{L} \left\| \mathcal{R}(F_l^T) - \mathcal{R}(F_l^S) \right\|_2^2,
\end{equation}

\noindent where $F^T$, $F^S$ are the teacher and student features from the $l$-th stage, respectively. The loss is computed across multiple stages to align global contextual representations at different scales. The hyperparameter $\lambda$ controls the contribution of the contextual alignment term. The contextual modeling module $\mathcal{R}(\cdot)$ is formulated as:

\begin{equation}
\mathcal{R}(F) = F + W_{v2} \cdot \text{ReLU} ( \text{GN} (
W_{v1} \cdot \sum_{j=1}^{N_v} \frac{e^{W_k F_j}}{\sum_{m=1}^{N_v} e^{W_k F_m}} \cdot F_j 
))
\end{equation}

\noindent where $F$ denotes the 3D feature map and $N_v = D \times H \times W$ is the number of voxels. The learnable parameters $W_k$, $W_{v1}$, and $W_{v2}$ are $1 \times 1 \times 1$ convolutional layers used for computing attention weights and feature transformations; $\text{GN}(\cdot)$ denotes group normalization. The inner summation implements a soft-attention aggregation of global contextual features, while the residual bottleneck refines the representation and preserves spatial semantics.

\vspace{-0.6em}
\subsection{Overall Loss}
To sum up, we define the overall objective as a combination of three components:
\begin{equation}
\mathcal{L}_{\text{total}} = \mathcal{L}_{\text{task}} + \mathcal{L}_{\text{MS-SARD}} + \mathcal{L}_{\text{MS-CA}}
\end{equation}
where $\mathcal{L}_{\text{task}}$ denotes the standard segmentation loss (e.g., Dice loss and cross-entropy) applied between the student prediction and the ground-truth labels.
\begin{table*}[th]
\centering
\caption{
Segmentation dice score for 13 abdominal organs, overall mean Dice (mDice), and HD95 (mm) on the BTCV dataset.
}
\resizebox{\textwidth}{!}{
\begin{tabular}{lccccccccccccccc}
\toprule
\textbf{Approach} & \textbf{Spl} & \textbf{RKid} & \textbf{LKid} & \textbf{Gall} & \textbf{Eso} & \textbf{Liv} & \textbf{Sto} & \textbf{Aor} & \textbf{IVC} & \textbf{Veins} & \textbf{Pan} & \textbf{Rad} & \textbf{Lad} & \textbf{mDice} & \textbf{HD95} \\
\midrule
SlimUNETR~\cite{pang2023slim} & 84.88 & 81.79 & 83.05 & 64.63 & 65.39 & 94.61 & 76.13 & 87.34 & 80.08 & 58.96 & 57.37 & 49.34 & 46.45 & 71.54 & 12.34\\
SegFormer3D~\cite{perera2024segformer3d} & 87.42 & 82.72 & 84.92 & 70.82 & 69.42 & 93.83 & 79.05 & 87.82 & 81.30 & 63.18 & 64.47 & 52.76 & 48.75 & 74.34 & 13.41 \\
\midrule
UNETR~\cite{hatamizadeh2022unetr} & 88.36 & 82.86 & 84.10 & 66.28 & 70.88 & 95.07 & 78.27 & 86.82 & 79.93 & 63.41 & 63.68 & 58.43 & 56.41 & 74.96 & 21.78 \\
nnU-Net~\cite{isensee2021nnu} & 88.00 & 84.78 & 85.47 & 71.85 & 75.44 & 95.32 & 76.54 & 90.93 & 85.14 & 68.91 & 62.19 & 66.90 & 60.19 & 77.82 & 10.91 \\
nnFormer~\cite{zhou2021nnformer} & 92.40 & 83.31 & 85.39 & 72.02 & 73.16 & 94.85 & 83.61 & 89.56 & 81.64 & 67.63 & 70.76 & 61.58 & 61.71 & 78.28 & 10.57 \\
TransBTS~\cite{wang2021transbts} & 89.92 & 84.31 & 85.59 & 73.86 & 72.09 & 96.20 & 81.49 & 89.81 & 85.23 & 65.67 & 71.20 & 63.74 & 66.15 & 78.87 & 14.47 \\
UNETR++~\cite{shaker2024unetr++} & 94.69 & 85.99 & 86.90 & 78.83 & 73.79 & 96.22 & 83.27 & 91.20 & 87.02 & 72.40 & 71.82 & 67.64 & 62.19 & 80.92 & 9.59 \\
MedNeXt~\cite{roy2023mednext} & 90.08 & 86.96 & 88.96 & 77.27 & 78.16 & 96.91 & 84.24 & 92.16 & 88.65 & 75.45 & 80.13 & 68.83 & 70.87 & 82.98 & 5.45\\
3D UX-Net~\cite{lee20223d} & 92.47 & 84.39 & 86.54 & 78.72 & 74.16 & 95.44 & 82.47 & 90.93 & 85.03 & 70.56 & 64.60 & 66.49 & 64.85 & 79.74 & 12.43 \\
SwinUNETR~\cite{tang2022self} & 88.56 & 85.92 & 86.03 & 79.40 & 75.50 & 95.41 & 79.58 & 90.13 & 86.18 & 71.12 & 69.36 & 69.35 & 65.19 & 80.13 & 14.01\\
SwinUNETRv2~\cite{he2023swinunetr} & 90.78 & 86.29 & 85.62 & 79.20 & 75.90 & 95.21 & 78.90 & 90.00 & 86.25 & 72.61 & 74.50 & 71.44 & 69.66 & 81.26 & 12.86\\
\midrule
\midrule
Teacher          & 96.39 & 94.72 & 94.92 & 73.84 & 80.32 & 97.33 & 87.16 & 90.68 & 87.82 & 75.21 & 84.51 & 72.94 & 77.40 & 85.64 & 4.66\\
Student baseline & 89.30 & 93.52 & 93.33 & 58.82 & 77.41 & 96.11 & 74.97 & 89.78 & 86.50 & 65.58 & 81.19 & 66.79 & 71.66 & 80.38 & 15.19\\
\midrule
\rowcolor{gray!20}
\textbf{Our ReCo-KD}  & 95.94 & 93.85 & 94.72 & 75.57 & 78.11 & 96.97 & 89.83 & 91.58 & 87.19 & 72.50 & 80.72 & 71.26 & 76.91 & 85.01 & 8.89\\
\bottomrule
\addlinespace[3pt]
\multicolumn{16}{p{1\linewidth}}{\footnotesize
\textit{Notes}: Teacher: nnU-Net with a residual encoder trained at full capacity. Student baseline: lightweight nnU-Net obtained by uniform channel scaling (\(t{=}2\), i.e., one-quarter of the original channels) without knowledge distillation. \textbf{Our ReCo-KD}: applies the proposed region- and context-aware distillation to the same lightweight student.
}\\
\end{tabular}
}
\vspace{-1em}
\label{tab:btcv-comparison}
\end{table*}

\vspace{-0.3em}
\section{Experiments and Results}

\subsection{Experimental Setups}
\subsubsection{Datasets} 

We evaluate ReCo-KD on four datasets: three public benchmarks and one private, more complex task.

\paragraph{BRATS 2021 Dataset \cite{baid2021rsna}}  Pre-operative multi-parametric MRI with four modalities (T1, T1Gd, T2, T2-FLAIR) and labels for Enhancing Tumor (ET), Peritumoral edema (ED) and necrotic–non-enhancing core (NCR/NET), typically evaluated as Whole Tumor (WT), Tumor Core (TC) and Enhancing Tumor (ET); the dataset contains 1251 cases, and we adopt an 80:20 split for training and validation.

\paragraph{MSD Hippocampus \cite{antonelli2022medical}}  Single-modality MRI with annotations for anterior and posterior hippocampus; 263 training and 131 test volumes.

\paragraph{BTCV \cite{landman2015miccai}} Abdominal CT with 13 organs annotated and 30 labeled training volumes. We use a 24/6 train/validation split.

\paragraph{Large-Scale Brain Structure Dataset (Private)}
A challenging fine-grained brain-structure dataset with 110 anatomical categories, including many small cortical and subcortical regions. It is aggregated from multiple public neuroimaging cohorts—ABIDE I \cite{di2014autism}, CoRR \cite{zuo2014open}, ADNI \cite{mueller2005ways}, and SALD \cite{wei2018structural}—yielding a total of 1,189 training subjects. Evaluation is performed on Mindboggle-101 \cite{klein2012101}, which provides cortical and subcortical segmentation that was completed by neuroimaging experts using manual delineation, ensuring anatomical accuracy and consistency of regional labeling.

\subsubsection{Implementation Details}
Our implementation builds on nnU-Net \cite{isensee2021nnu}. Unless otherwise stated, we use its default preprocessing/planning, deep supervision, data augmentation, and sliding-window inference.
The teacher is the nnU-Net model with the residual encoder \cite{isensee2024nnu}. The student is obtained by uniformly scaling the channel width with multipliers $\{1, \tfrac{1}{2}, \tfrac{1}{4}, \tfrac{1}{8}\}$ while keeping network depth, strides, patch size, and batch size unchanged. When channel dimensions differ during distillation, a $1\times 1\times 1$ adapter aligns features. For the distillation loss, we set the temperature to $0.5$ and the loss weights for activation–mask consistency ($\gamma$) and relation alignment ($\lambda$) are also fixed to $1$.  
Because every term is voxel- and channel-normalized, their magnitudes are naturally comparable. Other settings follow nnU-Net defaults. 

\paragraph{Cross-validation} We adopt nnU-Net’s five-fold protocol. Due to computational constraints, unless specified otherwise, we train and report results on a single fold, using the best-validation checkpoint within that fold.

\paragraph{Evaluation metrics} We report mean Dice (mDice), Normalized Surface Dice (NSD), and 95th-percentile Hausdorff distance (HD95),

\subsection{Main Results}

\begin{table}[ht]
\centering
\vspace{-1em}
\caption{
Segmentation Dice (\%, higher is better) on Hippocampus.
“Ant.” and “Post.” denote the anterior and posterior hippocampus.
}
\resizebox{0.48\textwidth}{!}{
\begin{tabular}{lccccc}
\toprule
\textbf{Approach} & \textbf{Ant.} & \textbf{Post.} & \textbf{mDice} & \textbf{Params} & \textbf{FLOPs} \\
\midrule
SlimUNETR~\cite{pang2023slim} & 87.19 & 85.38 & 86.29 & 1.79 & 20.17\\
SegFormer3D~\cite{perera2024segformer3d} & 87.44 & 85.48 & 86.46 & 4.50 & 5.03 \\
\midrule
UNETR~\cite{hatamizadeh2022unetr} & 88.01 & 86.34 & 87.18 & 92.78 & 82.60 \\
nnFormer~\cite{zhou2021nnformer} &  87.58 & 85.84 & 86.71 & 149.25 & 213.60 \\
TransBTS~\cite{wang2021transbts} & 88.39 & 86.68 & 87.54 & 31.58 & 110.34\\
UNETR++~\cite{shaker2024unetr++} & 88.51 & 87.01 & 87.76 & 42.62 & 53.99 \\
3D UX-Net~\cite{lee20223d} & 89.33 & 87.64 & 88.49 & 53.00 & 631.97\\
SwinUNETR~\cite{tang2022self} & 88.61 & 87.12 & 87.87 & 69.19 & 337.61 \\
SwinUNETRv2~\cite{he2023swinunetr} & 88.48 & 86.86 & 87.67 &83.19 &353.61 \\
\midrule
\midrule
Teacher & 89.82 & 88.16 & 89.00 & 100.22 & 569.02\\
Student Baseline & 88.66 & 86.79 & 87.72 &  1.57 & 9.17 \\ \midrule
\rowcolor{gray!20}
\textbf{Our ReCo-KD} & 89.70 & 88.15 & 88.93 & 1.57 & 9.17 \\
\bottomrule
\addlinespace[3pt]
\multicolumn{6}{p{251pt}}{\footnotesize
\textit{Notes}: Teacher is the full-capacity nnU-Net with a residual encoder, and the Student baseline is the same network with channel width scaled to \(t{=}3\) (\(\tfrac{1}{8}\) of the original channels) without distillation.
Parameter counts are in millions (M) and FLOPs (G) are measured on a single-channel \(96\!\times\!96\!\times\!96\) volume.}\\
\end{tabular}
}
\vspace{-1em}
\label{tab:hippocampus_results}
\end{table}

\begin{table}[ht]
\centering
\vspace{-1em}
\caption{
Segmentation dice score on Whole Tumor (WT), Enhancing Tumor (ET), Tumor Core (TC), overall mean Dice (mDice), and HD95 (mm) on the BraTS2021 dataset.
}
\resizebox{0.98\linewidth}{!}{
\begin{tabular}{lcccccc}
\toprule
\textbf{Approach} & \textbf{WT} & \textbf{ET} & \textbf{TC} & \textbf{mDice} & \textbf{HD95}\\
\midrule
TransVW \cite{haghighi2021transferable}        & 92.32 & 82.09 & 90.21 & 88.21 & 8.33\\
UNet3D \cite{ronneberger2015u}         & 92.69 & 84.10 & 87.10 & 87.93 & 6.42\\
E1D3 UNet \cite{bukhari2021e1d3}      & 92.42 & 82.16 & 86.53 & 87.13 & 8.18\\
VNet \cite{milletari2016v}           & 91.38 & 86.90 & 89.01 & 89.09 & 9.83\\
nn-UNet \cite{isensee2021nnu}        & 92.71 & 88.34 & 91.39 & 90.84 & 5.33 \\
SegResNet \cite{myronenko20183d}      & 92.73 & 88.31 & 91.31 & 90.78 & 5.17 \\
AttUNet \cite{oktay2018attention}        & 92.02 & 88.28 & 90.94 & 90.40 & 6.02\\
\midrule
Swin UNETR \cite{hatamizadeh2021swin}     & 93.32 & 89.08 & 91.69 & 91.36 & 5.03\\
TransBTS \cite{wang2021transbts}       & 91.05 & 86.75 & 89.76 & 89.18 & 6.72\\
TransUNet \cite{chen2021transunet}      & 87.68 & 83.34 & 82.75 & 84.59 & 10.02\\
UNETR \cite{hatamizadeh2022unetr}           & 92.53 & 87.59 & 90.78 & 90.31 & 6.13\\
UNETR++ \cite{shaker2024unetr++}         & 91.62 & 86.35 & 92.17 & 90.05 & 6.17\\
VitAutoEnc \cite{cardoso2022monai}     & 81.41 & 68.35 & 78.66 & 76.14 & 17.92\\
VIT3D \cite{dosovitskiy2020image}          & 53.86 & 41.16 & 64.89 & 53.31 & 29.07\\
\midrule
\midrule
Teacher & 93.88 & 88.53 & 92.50 & 91.65 & 3.69\\
Student Baseline & 92.78 & 85.17 & 90.87 & 89.55 & 5.17 \\
\midrule
\rowcolor{gray!20}
\textbf{Our ReCo-KD} & 93.71 & 87.38 & 92.20 & 91.09 & 3.73\\
\bottomrule
\addlinespace[3pt]
\multicolumn{6}{p{0.92\linewidth}}{\footnotesize
\textit{Notes}: Teacher is the full-capacity nnU-Net with a residual encoder, and the Student baseline is the same network with channel width scaled to \(t{=}2\) (\(\tfrac{1}{4}\) of the original channels) without knowledge distillation.}\\
\end{tabular}
}
\vspace{-1em}
\label{tab:brats_results}
\end{table}

\subsubsection{BTCV Dataset Results}
\label{sec:btcv_results}
Table~\ref{tab:btcv-comparison} reports Dice across 13 abdominal organs. We compare against CNN methods (e.g., MedNeXt) and Transformer backbones (e.g., SwinUNETRv2).
Our distilled student achieves the best mean Dice 85.01\%, surpassing SwinUNETRv2 (81.26\%) and MedNeXt (82.98\%). Gains are largest on small/rare structures (pancreas, adrenal glands, gallbladder), alleviating the voxel-imbalance failures of the non-distilled student. Performance on large organs (e.g., liver, spleen) remains strong.

\subsubsection{Hippocampus Dataset Results}
\label{sec:hippocampus_results}
Table~\ref{tab:hippocampus_results} shows Dice (\%). Our method attains a mean Dice 88.93\%, exceeding SwinUNETRv2 (87.67\%) and matching or surpassing other lightweight models (e.g., SlimUNETR). This performance is obtained with only 1.57 M parameters and 9.17 GFLOPs at an aggressive channel-reduction setting of t=3 ($\frac{1}{8}$ of the teacher’s channel), demonstrating an excellent balance between accuracy and computational efficiency.

\subsubsection{BraTS 2021 Dataset Results}
\label{sec:brats_results}

As summarized in Table~\ref{tab:brats_results}, our method achieves average Dice 91.09\%, close to the teacher (91.65\%). The largest improvement is on ET, with +2.21 Dice over the non-distilled student, indicating better delineation of small enhancing lesions. WT and TC remain stable.

\subsubsection{Evaluation on a Large-Scale Brain Structure Dataset with 110 Categories}
\begin{table}[t]
\centering
\caption{Performance–efficiency trade-off on Brain Structure Segmentation.
}
\label{tab:brain_structure}
\setlength{\tabcolsep}{3.5pt}
{\footnotesize
\begin{tabular*}{\columnwidth}{@{\extracolsep{\fill}}lcrrrr@{}}
\toprule
Model & $t$ & P [M] & F [G] & D [\%] & T (CPU/2080Ti/H100) [s] \\
\midrule
Teacher            & 0 & 102.44 & 3364.88 & 81.65 & 119.00 / 2.07 / 1.02 \\ \midrule
Student (Base)     & 1 &  25.64 &  853.28 & 79.48 & 58.19 / 1.45 / 0.63 \\
\rowcolor{gray!20}
\textbf{Our ReCo-KD}   & 1 &  25.64 &  853.28 & 81.06 & 58.19 / 1.45 / 0.63 \\ \midrule
Student (Base)     & 2 &   6.43 &  219.35 & 78.92 & 34.62 / 1.16 / 0.42 \\
\rowcolor{gray!20}
\textbf{Our ReCo-KD}  & 2 &   6.43 &  219.35 & 80.25 & 34.62 / 1.16 / 0.42 \\
\bottomrule
\addlinespace[3pt]
\multicolumn{6}{p{0.95\linewidth}}{\footnotesize
\textit{Notes}: P [M] = parameter count (millions); F [G] = FLOPs (billions) for a 1×128×128×128 single-channel volume; D [\%] = mDice; Time [s] = mean per case over 101 test cases on (CPU/2080Ti/H100). \(t\): channels reduction factor.}\\
\end{tabular*}
}
\end{table}

Table \ref{tab:brain_structure} summarizes parameter counts, FLOPs, accuracy, and CPU/GPU inference times on a more complex task. With a quarter of the teacher channels (t = 2), our ReCo-KD trained student model retains 98.29\% of the teacher’s accuracy while reducing parameters by 93.72\% and FLOPs by 93.48\%. The CPU inference time decreases from 119 s to 34.6 s—a 70.92\% reduction. These results demonstrate that ReCo-KD maintains high accuracy under aggressive model compression for complex, fine-grained brain-region segmentation, supporting deployment in resource-constrained settings.

\begin{table}[t]
  \centering
  \vspace{-1em}
  \caption{Comparison of knowledge distillation methods on BTCV and BraTS2021. Reported metrics are Dice and Normalized Surface Dice (NSD), higher is better.}
  \label{tab:btcv-kd-comparison}
  \resizebox{0.98\linewidth}{!}{
  \begin{tabular}{lcccc}
    \toprule
    & \multicolumn{2}{c}{\textbf{BTCV}} & \multicolumn{2}{c}{\textbf{BraTS2021}} \\
    \cmidrule(lr){2-3}\cmidrule(lr){4-5}
    \textbf{Model} & \textbf{mDice (\%)} & \textbf{NSD (\%)} & \textbf{mDice (\%)} & \textbf{NSD (\%)} \\
    \midrule
    Teacher & 85.64 & 85.55 & 91.65 & 93.82 \\
    \midrule
    Student (w/o KD) & 80.38 & 78.53 & 89.55 & 90.64 \\
    SKD \cite{liu2023SKD} & \underline{84.53} & \underline{84.06} & 90.12 & 92.15 \\
    CWD \cite{shu2021channel} & 84.44 & 83.94 & \underline{90.99} & \underline{93.19} \\
    IFVD \cite{wang2020intra} & 84.28 & 83.32 & 89.80 & 92.89 \\
    FitNet \cite{romero2015fitnetshintsdeepnets} & 82.86 & 82.59 & 88.70 & 91.68 \\
    AT \cite{zagoruyko2016paying} & 82.18 & 80.94 & 90.17 & 92.38 \\
    CIRKD \cite{yang2022cross} & 81.91 & 80.77 & 90.62 & 93.09 \\
    \midrule
    \rowcolor{gray!20}
    \textbf{Our ReCo-KD} & \textbf{85.01} & \textbf{84.29} & \textbf{91.09} & \textbf{93.83} \\
    \bottomrule
    \addlinespace[3pt]
    \multicolumn{5}{p{245pt}}{\footnotesize
    \textit{Notes}: The teacher is the full-capacity nnU-Net with a residual encoder, and all student models (including ours) use uniform channel reduction to one-quarter of the teacher' channel ($t=2$).}\\
  \end{tabular}
  }
\vspace{-1em}
\end{table}

\vspace{-0.6em}
\subsection{Comparison with Other Knowledge Distillation Methods}
To further validate the effectiveness of ReCo-KD, we compare it with general-purpose and segmentation-specific knowledge distillation (KD) approaches. General-purpose baselines include FitNet~\cite{romero2015fitnetshintsdeepnets} and AT~\cite{zagoruyko2016paying}. Segmentation-oriented methods include SKD~\cite{liu2023structured}, IFVD~\cite{wang2020intra}, CWD~\cite{shu2021channel}, and CIRKD~\cite{yang2022cross}. Results on BTCV and BraTS2021 are summarized in Table~\ref{tab:btcv-kd-comparison}.
On BTCV, ReCo-KD achieves a Dice of 85.01\% and NSD of 84.29\%, on par with the best existing methods (e.g., SKD).
On BraTS2021, ReCo-KD attains 91.09\% Dice and 93.83\% NSD, reaching state-of-the-art performance among the compared KD methods.
\vspace{-0.6em}
\subsection{Qualitative Analysis}

\subsubsection{Comparison with Ground Truth, Teacher, and Student Baseline}

\begin{figure}[ht]
  \includegraphics[width=0.48 \textwidth]{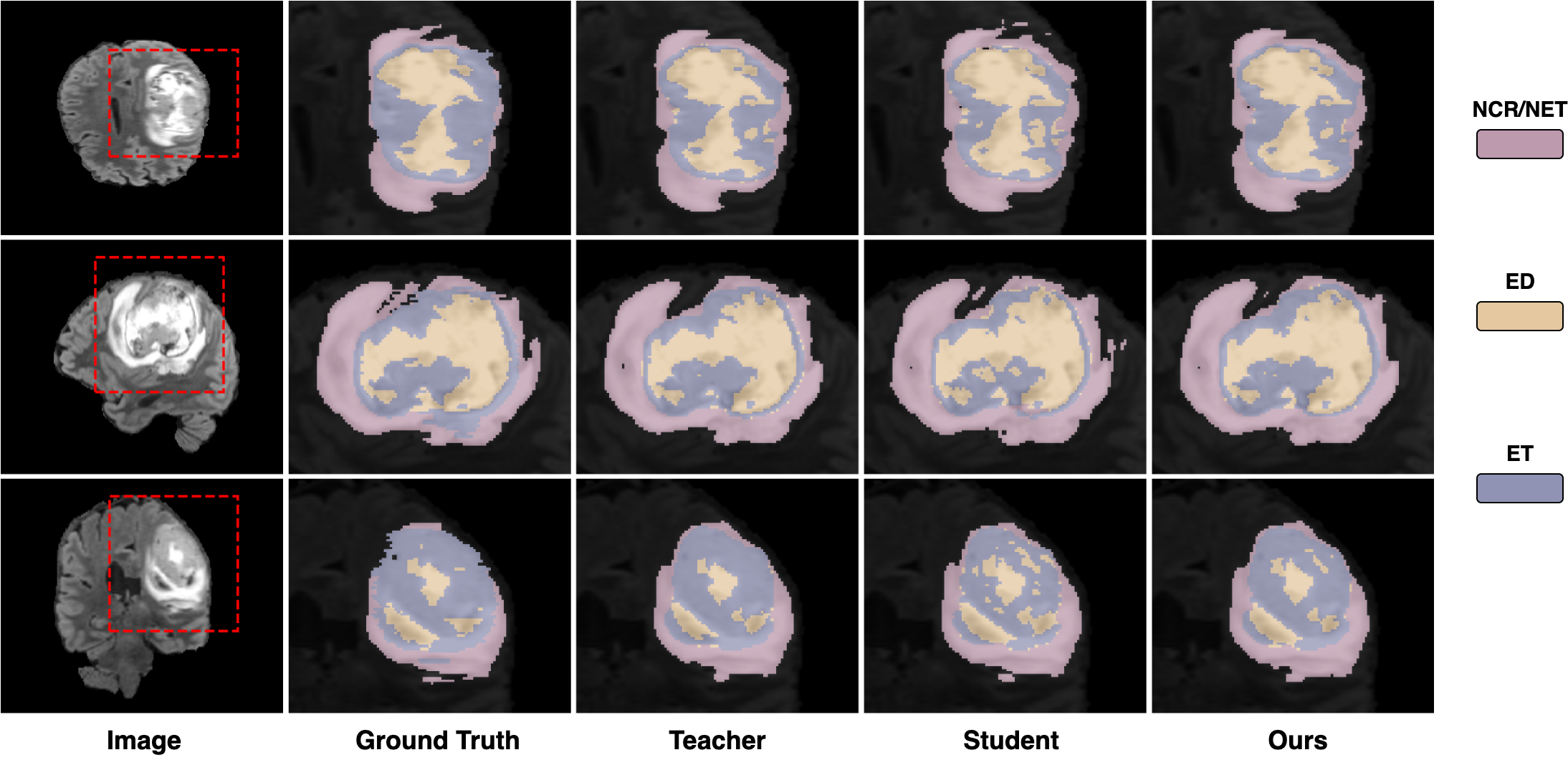}
  \caption{Qualitative results on BraTS2021. Rows show axial, sagittal, and coronal views. The first column is the full slice with a red dashed box marking the region of interest; the others show the cropped region for Ground Truth, Teacher, Student, and our ReCo-KD. Zoom for the best view.}
  \label{fig:brats2021_reco_kd}
\end{figure}

Fig.~\ref{fig:brats2021_reco_kd} shows BraTS2021 qualitative results (axial, sagittal, coronal) for Ground Truth, Teacher, Student, and Ours. The non-distilled Student tends to under-segment ET and produce irregular boundaries near the core. In contrast, ReCo-KD yields crisper boundaries and better overlap across subregions, indicating effective transfer of structural cues from the Teacher and mitigation of Student artifacts.

\subsubsection{Comparison with Other Knowledge Distillation Methods}

\begin{figure*}[!t]
  \centering
  \includegraphics[width=\textwidth]{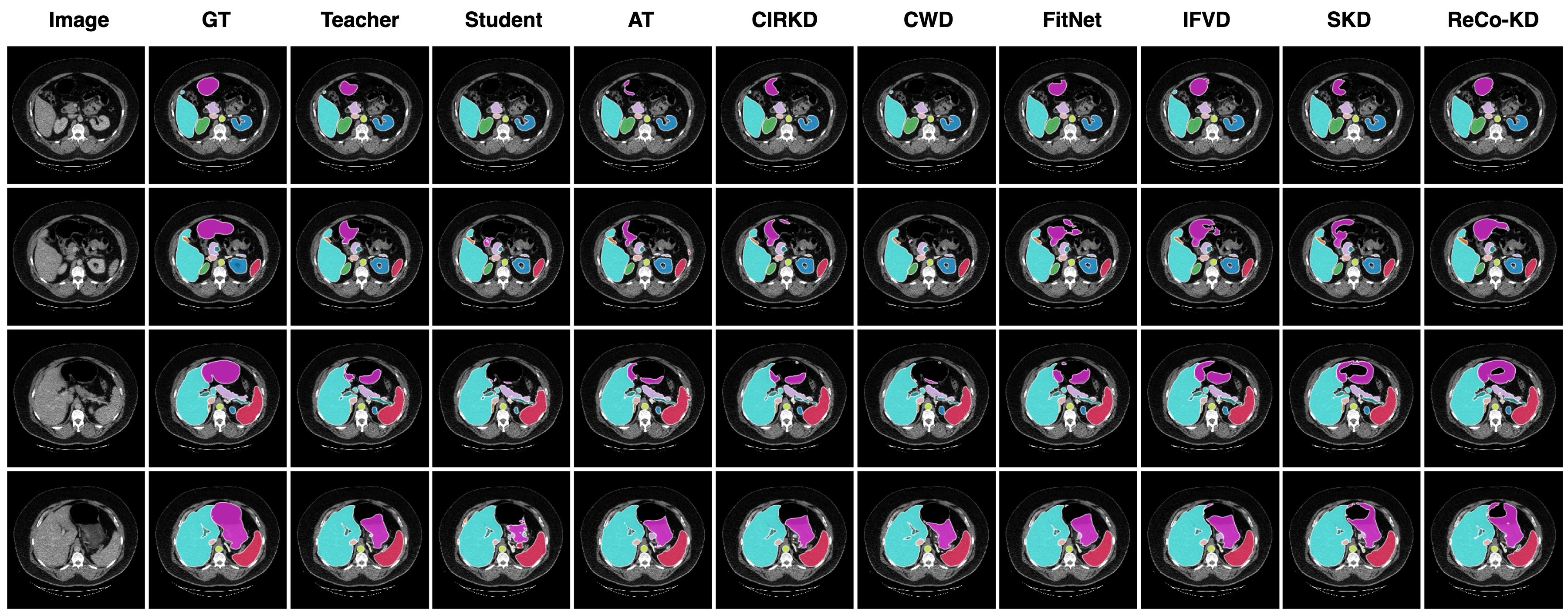}
  \caption{Qualitative comparison on BTCV. For each axial slice (row), the first column shows the CT image, the second shows the ground truth, and the remaining columns depict predictions of different methods; our method is fixed at the far right of each row.}
  \label{fig:btcv_benchmark_vis}
\end{figure*}

We further compare with representative KD methods on BTCV (see Fig. \ref{fig:btcv_benchmark_vis}). All visualizations adopt identical windowing and color mapping for fairness. Our KD method (ReCo-KD) shows crisper organ boundaries and fewer spurious regions.

\vspace{-0.6em}
\subsection{Ablation Studies}
We ablate ReCo-KD on BTCV to isolate the effect of each component and setting.

\subsubsection{Effect of Distillation Components}
Using the non-distilled Student as the baseline, Table~\ref{tab:btcv-ablation} reports Dice, NSD, and HD95 for each variant.
\emph{FG-distill} applies the region loss only within the teacher’s foreground mask, \emph{BG-distill} applies it only on background voxels, and \emph{Mask-align} removes the region loss while aligning teacher–student activation masks across scales; \emph{MS-CA only} performs multi-scale contextual alignment without any region loss. 
All single-component settings improve Dice and NSD over baseline and reduce HD95. Among MS-SARD variants, \emph{FG-distill} is strongest; MS-CA only also yields consistent gains (+2.00\% Dice). Our \emph{ReCo-KD} (MS-SARD + MS-CA) attains the best overall results, evidencing complementarity between region cues and multi-scale context.

\begin{table}[h]
  \centering
  \vspace{-1em}
  \caption{Ablation study of different components on BTCV. Best and second-best are in bold and underlined. 
  } 
  \label{tab:btcv-ablation}
  \resizebox{0.98\linewidth}{!}{
  \begin{tabular}{lcccc}
    \toprule
    \textbf{Setting} & \textbf{mDice} & \textbf{NSD} & \textbf{HD95} & $\Delta$ mDice \\
    \midrule
    Student (w/o KD)         & 80.38 & 78.53 & 15.23 & -- \\
    \midrule
    MS-SARD: Mask-align      & 82.44 & 80.31 &  6.53 & +2.06 \\
    MS-SARD: FG-distill      & \uline{83.61} & \uline{82.72} & \uline{6.16} & +3.23 \\
    MS-SARD: BG-distill      & 83.20 & 81.71 & 10.24 & +2.82 \\
    MS-CA only               & 82.38 & 81.00 & 11.91 & +2.00 \\
    \midrule
    ReCo-KD: MS-SARD + MS-CA & \textbf{85.01} & \textbf{82.41} & \textbf{6.10} & +4.63 \\
    \bottomrule
    \addlinespace[3pt]
    \multicolumn{5}{p{0.95\linewidth}}{\footnotesize
    \textit{Notes}: Mask-align = activation-mask alignment only (no region loss); FG-distill = foreground-only region distillation; BG-distill = background-only region distillation; MS-CA = contextual alignment only. $\Delta$Dice is relative to Student (w/o KD).
    }\\
  \end{tabular}
  }
  \vspace{-1em}
\end{table}

\begin{table}[t]
  \centering
  \vspace{-1em}
  \caption{Ablation of width scaling on BTCV using $t$, where channels are multiplied by $2^{-t}$.
  }
  \label{tab:ablation_reduction_btcv}
  \begin{tabular}{lccccc}
    \toprule
      $t$ \;($\times 2^{-t}C$) &
      \multicolumn{1}{c}{\shortstack{Params\\(M) $\downarrow$}} &
      \multicolumn{1}{c}{\shortstack{FLOPs\\(G) $\downarrow$}} &
      \multicolumn{1}{c}{\shortstack{Max Mem\\(GB) $\downarrow$}} &
      \multicolumn{1}{c}{\shortstack{Inf.\ Time\\(s) $\downarrow$}} &
      \multicolumn{1}{c}{\shortstack{mDice \\ (\%) $\uparrow$}} \\
    \midrule
    $t=0$ ($\times 1$)     & 141.41 & 2066.65 & 12.43 & 10.38 & 85.64 \\
    $t=1$ ($\times \tfrac{1}{2}$) & 35.37  & 518.16  & 6.13  & 5.00  & 85.11 \\
    $t=2$ ($\times \tfrac{1}{4}$) & 8.85   & 130.29  & 3.09  & 3.40  & 85.01 \\
    $t=3$ ($\times \tfrac{1}{8}$) & 2.22   & 32.95   & 1.64  & 2.70  & 80.96 \\
    \bottomrule
    \addlinespace[3pt]
    \multicolumn{6}{p{0.95\linewidth}}{\footnotesize
    \textit{Notes}: FLOPs @ $128^3$; inference time = mean per-case over BTCV ($n{=}30$) with native shapes, measured on a single RTX 2080 Ti with AMP.}\\
  \end{tabular}
  \vspace{-1em}
\end{table}

\subsubsection{Efficiency Analysis of Channel Reduction Factor}

As shown in Table~\ref{tab:ablation_reduction_btcv}, we ablate channel width scaling via the factor $2^{-t}$. Under uniform width changes for both encoder and decoder, parameters and FLOPs are expected to scale as $2^{-2t}$ and peak memory as $2^{-t}$; the empirical results closely match this trend. Relative to $t=0$, FLOPs drop by $ 93.7\%$ while peak memory decreases by $67.2\%$ at $t=2$. Inference latency improves from 10.38s to 3.40s (about $3.05\times$ faster). Accuracy remains effectively unchanged through $t=2$ (mDice 85.64\% to 85.01\%), but degrades at $t=3$. We therefore adopt $t=2$ as the default trade-off, consider $t=1$ when accuracy is paramount, and reserve $t=3$ for strict resource budgets.
\begin{table}[h]
\centering
\vspace{-1em}
\caption{Feature distillation at different \emph{encoder} stages on BTCV (student tested at $t{=}2$).}
\label{tab:btcvs_kd_stage}
\resizebox{\linewidth}{!}{%
\begin{tabular}{lcccc}
\toprule
\textbf{Enc.\ stages} & \textbf{mDice (\%)} & \textbf{NSD (\%)} & \textbf{HD95 (mm)} & {$\Delta$mDice} \\
\midrule
none    & 80.38 & 78.53 & 15.19 & -- \\
0-{}-1    & 82.08 & 80.03 & 13.86 & +1.70 \\
2-{}-3    & 82.94 & 81.25 & 10.90 & +2.56 \\
4-{}-5    & 83.35 & 81.98 &  9.12 & +2.97 \\
\textbf{Our ReCo-KD (0-{}-5)}    & \textbf{85.01} & \textbf{84.29} & \textbf{8.89} & \textbf{+4.63} \\
\bottomrule
\end{tabular}
}
\vspace{-1em}
\end{table}

\subsubsection{Effect of encoder-stage choices.}

Table \ref{tab:btcvs_kd_stage} compares feature-distillation across different combinations of encoder stages, from shallow to deep. As the distilled stages move deeper, Dice and NSD steadily increase while HD95 decreases, indicating that high-level semantic features provide stronger guidance than low-level details. Distilling from all stages achieves the best performance, confirming that multi-scale supervision—combining fine spatial cues with rich semantic context—offers the most comprehensive benefit for the student model.
\section{Limitations and Future Work}
This study evaluates ReCo-KD using the default nnU-Net student and a homogeneous CNN-to-CNN setting, which restricts the architectural search space and may understate the benefits of stronger lightweight students. Future work will explore diverse student designs within nnU-Net—including depth/width scaling, alternative encoders/decoders, and efficient attention—and investigate heterogeneous distillation between CNN and Transformer backbones.
\section{Conclusion}


We introduced ReCo-KD, a region- and context-aware knowledge distillation framework for 3D medical image segmentation. By combining multi-scale structure-aware region distillation with multi-scale contextual alignment, the method effectively transfers both fine anatomical details and long-range contextual dependencies from a high-capacity teacher to a lightweight student.
Built on the self-configuring nnU-Net pipeline, ReCo-KD requires no custom student design and is easy to integrate into existing workflows. Extensive experiments on BTCV, BraTS2021, Hippocampus, and a large-scale 101-region brain dataset show that ReCo-KD consistently narrows the teacher–student performance gap while cutting parameters and FLOPs by up to 94\% and 93\%, respectively, and reducing CPU inference time by more than 70\%. These results demonstrate that ReCo-KD enables accurate, resource-efficient deployment of 3D segmentation models in real clinical settings.

\bibliographystyle{IEEEtran}
\bibliography{ref}

\end{document}